\UseRawInputEncoding 
\documentclass{article}

\usepackage{enumerate}
\usepackage{amsthm,amsmath}
\usepackage{amssymb,color} 
\usepackage[margin=1in]{geometry}
\usepackage{graphicx}
\usepackage{url}

\usepackage[round]{natbib}
\newtheorem{theorem}{Theorem}[section]
\newcommand{\citem}[1]{\citeauthor{#1} (\citeyear{#1})}

\def\R{{\mathbb R}}

\def\E{{{\mathbb E}\,}}

\date{}

\begin{document}
\begin{center}
	{\LARGE A Statistician Teaches Deep Learning}
	
	\bigskip
	
	G. Jogesh Babu, Pennsylvania State University\\
	David Banks, Duke University\\
	Hyunsoon Cho, National Cancer Center of Korea\\
	David Han, University of Texas San Antonio\\
	Hailin Sang, University of Mississippi\\
	Shouyi Wang, University of Texas Arlington
	
	\bigskip

\end{center}

\bigskip

\begin{center}
	\bigskip
	
\textbf{Abstract}
\end{center}	
Deep learning (DL) has gained much attention and become increasingly popular in modern data science. Computer scientists led the way in developing deep learning techniques, so the ideas and perspectives can seem alien to statisticians.
Nonetheless, it is important that statisticians become involved---many of our students
need this expertise for their careers.
In this paper, developed as part of a program on DL held at the Statistical
and Applied Mathematical Sciences Institute, we address this culture gap and provide tips on how to teach deep learning to statistics graduate students. 
After some background, we list ways in which DL and statistical
perspectives differ, provide a recommended syllabus that evolved from teaching
two iterations of a DL graduate course, offer examples of suggested homework
assignments, give an annotated list of teaching resources, and discuss DL in the
context of two research areas.


\bigskip

\section{Introduction}

Deep learning (DL) has become an essential methodology in many industries, but it is not yet part of the standard statistical curriculum.
We believe it is important to bring DL into the purview of statisticians.  But there are challenges---aspects of DL do not align with 
conventional statistical wisdom, and there are practical and theoretical barriers that arise when trying to teach this material.

Nonetheless, DL has achieved tremendous successes in the last decade which make it a critical statistical tool and object of study. It is the foundation for autonomous vehicles (\citem{gt2020}, \citem{lc2017}), which have the potential for revolutionary advances in safety, fuel economy, and congestion reduction (\citem{ma2015}, \citem{pp2017}, \citem{rl2017}, \citem{ws2015}).
DL can read x-rays more accurately than trained radiologists (\citem{ai2019}, \citem{rn2018}, \citem{zj2019}), it outperforms humans in complex strategy games (\citem{sp2015}, \citem{vb2019}), and has set the standard for automatic language translation (\citem{lm2016}, \citem{lp2015}, \citem{sl2016}). 
Looking ahead, DL is poised to control robotic sorting of waste (\citem{ba2018}, \citem{sm2016}), coordinate complex medical treatment and reimbursement management in the U.S.\ healthcare system (\citem{am2020}, \citem{dw2108}, \citem{lt2017}), and optimize dynamical systems in high-tech manufacturing (\citem{sy2019}). DL techniques are especially popular in image classification problems, image synthesis, and speech recognition (\citem{cp2018}, \citem{gw2017}, \citem{hy2018}, \citem{i2018}, \citem{kk2018}, \citem{ny2015}, \citem{nt2018}, \citem{s2016}, \citem{ss2015}, \citem{wv2015}, \citem{zy2018}).
 
To remain viable and relevant, statisticians must master DL.  
That goal means that some of us need to learn how to teach DL.  
But there are practical, pedagogic, and technical difficulties.  
This paper will address all three hurdles.

A DL course needs to balance practice and theory, but (for now) the main emphasis should be practice.
There is a growing body of DL theory that we could and probably should teach, but the appetite and drive among the students is to do DL, 
not to prove things about DL.  
But statisticians generally teach theory before practice, and so we need to adapt.  
One result of that choice is that there are many platforms for DL programming, and it is problematic to manage a class in which students are 
programming in different environments.

Our advice is to either let students use whatever software tools they choose, or to enforce a single environment---if the students are
from the same department and have a programming lingua franca, we recommend the latter.
But in many situations that homogeneity is unrealistic---graduate courses in DL are popular and attract students from many different
disciplines with many different backgrounds.
  
The major open source environments for DL are TensorFlow, Keras, and PyTorch, but it is a fluid situation and other contenders include
Caffe, theano, mxnet (Apache), CNTK (Microsoft), DL4J, Chainer, and fast.ai.  
When picking an environment, one should consider the learning curve, the rate of new development, the commitment of the user 
community, the likelihood it will be stable and sustained, and the ecosystem of tools.
How one weights these different aspects depends upon one's purpose, applications, and background.

If we had to choose, we would select TensorFlow, partly because it is strongly supported by Google (\citem{h2018}, \citem{l2019}).  The Google Brain team developed it, and they designed it to take advantage of Google's powerful tensor processing units, or TPUs.  TensorFlow has a vigorous community of open-source developers, notably H2O, and it has the top ratings from \citem{zb2018} and from Hale (\url{http:/bit.ly/2GBa3tU}).  Keras and PyTorch are both good choices, and PyTorch seems to be gaining in popularity faster than Keras, but we have no specific data that confirms that impression.
PyTorch was developed at Facebook, and is also open source; its Python interface is popular.  PyTorch is the basis for the Tesla Autopilot (\citem{s2019}) and Uber's Pyro (\citem{po2019}).  Keras is an open source DL library that can run on top of TensorFlow, and it is also supported by Google.
Keras is written in Python, and was originally conceived as an interface rather than a standalone DL platform---the intent was to provide high-level abstractions
that enable developers to work fairly independently of the underlying software system.
But those high-level abstractions may obscure important pedagogical lessons by sweeping aside practical complexities and enforcing certain
default choices.

The main problem with allowing students to use whatever software they prefer is that it becomes very difficult to compare their performance 
on class assignments. 
In this kind of course it is important for students to learn which techniques work well in which situations, and the success of those
techniques can be confounded with the choice of software environment.
Performance also depends upon how long the students train their deep networks, the initialization of the parameters, and numerous technical choices in how
the DL training is done.
In Section 3.2 we provide examples of model homework assignments we constructed to put evaluations on a common footing.

Finding a good textbook is another practical problem.  
The bible in the field is {\em Deep Learning} by 
\citem{gb2016}.  
But the book was not written for statisticians, and we find it presents cognitive headwinds.
For example, we are concerned about the bias-variance tradeoff and the Curse of Dimensionality, which are only lightly touched upon in the text.
So our advice is that statistical educators write standalone lecture notes, 
which may draw heavily upon \citem{gb2016}, but which also employ other sources.
One reason is that the first five chapters 
introduce all the mathematical technicalities (e.g., the Moore-Penrose inverse, 
principal components analysis, overflow and underflow, maximum likelihood, Bayesian inference, and tensors).  
Later on, when these concepts are used, the text assumes that the reader recalls 
that material flawlessly.  
In our experience, it is better for mathematical concepts be introduced on a just-in-time basis.

Statisticians also face a terminology barrier.
For example, the hidden nodes in a deep neural network have activation functions, which are often logistic-like functions of a linear transformation 
of the data with the familiar form $\beta_0 + \bf{x}^{\prime} \bf{\beta}$.  
In the DL community, $\beta_0$ is called a bias.  
Other terms that are not used in the way a statistician expects are noise robustness, normalization, and hyperparameter. 
These are minor points, but they can cause confusion.

In this paper, we advise on how to teach DL from a statistical perspective. 
Further, we urge our colleagues to bring DL into our graduate programs---the students want and need this education.
Section 2 lays out ways in which the computer scientist's view and the statistician's view do not align.
Section 3 provides a detailed syllabus and two model assignments.
Section 4 lists more educational resources, and Section 5 concludes.

\section{Different Worldviews}

Statisticians have different training and instincts than computer scientists.
This section lists some of the disjunctions that became evident when teaching 
a DL graduate course in statistics.

\subsection{Function Composition}
One contrast between statisticians and DL computer scientists is that we generally make predictions and classifications
using linear combinations of basis elements---this is the cornerstone of nearly all regression and much classification,
including support vector machines, boosting, bagging, and stacking (nearest-neighbor methods are a rare
exception).
But DL uses compositions of activation functions of weighted linear combinations of inputs.
Function composition enables one to train the network through the chain rule for derivatives, which is the 
heart of the backpropagation algorithm.

The statistical approach generally enables more interpretability, but function composition increases
the space of models that can be fit.
If the activation functions in the DL networks were linear, then DL in a regression application would
be equivalent to multiple linear regression, and DL in a classification application would be equivalent
to linear discriminant analysis.
In that same spirit, a deep variational autoencoder with one hidden layer and linear activation functions
is equivalent to principal components analysis.
But use of nonlinear activation functions enables much more flexible models.

Statisticians are taught to be wary of highly flexible models---these have the potential to 
interpolate the training data, leading to poor generalizability.
Recent work indicates that DL actually does interpolate the training data, but
still performs well on test data (\citem{BHMM}).
 
\subsection{Parameterization}

Statisticians worry about overparameterization and issues related to the
Curse of Dimensionality (COD), as in \citem{Hastie2009}.  But DL frequently has millions of parameters.
Even though DL typically trains on very large datasets, their size still cannot
provide sufficient information to overcome the COD.

Statisticians often try to address overparameterization through variable selection,
removing terms that do not contribute significantly to the inference.
This generally improves interpretability, but often makes relatively strong use
of modeling assumptions.

In contrast, DL wants all the nodes in the network to contribute only slightly
to the final output, and uses dropout (and other methods) to ensure that result.
Part of the justification for that is a biological metaphor---genome-wide association
studies find that most traits are controlled
by many, many genes, each of which has only small effect.

And, of course, with the exception of \citem{chen2019looks}, DL typically is not concerned
with interpretability, which is generally defined as the ability to identify the 
reason that a particular input leads to a specific output.
DL does devote some attention to explainability, which concerns the extent
to which the parameters in deep networks can shape the output (\citem{samek2017explainable}).
To statisticians, this is rather like assessing variable importance.

\subsection{Theory}

Statistics has its roots in mathematics, so theory has primacy in our community.
Computer science comes from more of a laboratory science tradition, so the emphasis is upon performance.

This is not say that there is no theory behind DL; it is being built out swiftly.
But much of it is done after some technique has been found to work, rather than
as a guide to discovering new methodology.

One of the major theorems in DL concerns the universal approximation property.
Let $\mathcal{F}$ be the class of functions generated by a network with layer width $m$ and non-constant, bounded, and continuous  activation functions $\sigma$. 
This network has the universal approximation property if for any $\epsilon>0$ and any continuous function 
$f(x)$ on $[0,1]^d$, there exists an integer $m=m(f, \epsilon)$, such that 
$$\inf_{g\in \mathcal{F}}\|f-g\|_{L^\infty([0,1]^d)}\le \epsilon.$$
The following result is foundational in DL.

\begin{theorem}
Neural networks with smooth activation functions that are not polynomials have the universal approximation property. 
\end{theorem}
\citem{Cybenko} and \citem{HSW} independently proved an early version of this theorem for shallow 
feedforward neural networks with sigmoid activation functions. 
It was later shown in \citem{LLPS} that the class of deep neural networks is a universal approximator if and only if the activation function is not polynomial.

\subsection{Bounds On Generalization Error}

One of the most important theoretical results for DL are upper bounds on its generalization error (also known as expected loss). 
Such bounds indicate how accurately DL is able to predict outcomes for previously unseen data. 
Sometimes the bounds are unhelpfully large, but in other cases they are 
usefully small.

The generalization error for any function $f(x)$ is the expectation $L(f)=\E [\ell(f(X),Y)]$, where the function $\ell$ is the loss function (e.g., squared error for continuous outputs or the indicator loss function for classification). An objective of machine learning is to develop an algorithm to find functions $\hat{f}_n(x)=\hat{f}_n(x; X_1, Y_1, \cdots, X_n,Y_n)$ based on the training data set $\{(X_i,Y_i)\}, X\in \R^d,Y\in \R$ that predict well on unseen data.  
The standard approach to evaluating the generalization error $L(\hat{f}_n)$ of the algorithm function $\hat{f}_n(x)$ is to study the quantity $\sup_{f\in \mathcal{G}}|L(f)-\hat{L}(f)|$ for a suitable class of functions $\mathcal{G}$. Here $\hat{L}(f)=\frac{1}{n}\sum_{i=1}^n\ell(f(X_i), Y_i)$ is the empirical risk of a function $f$.  One of the classical metrics which is used to evaluate $\sup_{f\in \mathcal{G}}|L(f)-\hat{L}(f)|$ and hence the generalization error of learning algorithms is the Rademacher complexity.
Rademacher complexity for generalization error bounds has been studied by many researchers (\citem{Koltchinskii2}, \citem{Koltchinskii1}, \citem{BBL}, \citem{BFT}, \citem{GRS}, \citem{XW}). 

Other new theoretical results about DL include Vapnik-Cervonenkis bounds which partially address concerns previously raised about overly flexible models
(\citem{shalev2014understanding}).
And\\ \citem{betancourt2018symplectic} use symplectic geometry and dynamical systems
theory to set hard limits on the rate at which a DL network
can be trained.

\subsection{The Double Descent Curve}

In statistics, we worry about the bias-variance tradeoff, and know that overtraining causes performance on the test sample to deteriorate after
a certain point, so the error rate on the test data is U-shaped as a function
of the amount of training.
But in DL, there is a double descent curve; the test error decreases,
rises, then decreases again, possibly finding a lower minimum before
increasing again.  This is thought to be because the DL network
is learning to interpolate the data.  

The double descent phenomenon is one of the most striking results in modern 
DL research in recent years. It occurs in CNNs, ResNets, and transformers: as the number of parameters increases in a neural network, the test error initially decreases, then increases, and, just as the model approaches interpolation with approximately zero training error, the test error starts a second descent. The first switch happens in the under-parameterized regime which is consistent with the classical bias/variance tradeoff in the standard statistical machine learning theory. In the over-parameterized regime, where the model complexity is large compared to the sample size, DL practitioners often claim that Òbigger models are betterÓ; c.f.,  \citem{Krizhevsky}, \citem{Szegedy}, \citem{hc2019}.  This model-wise double descent was first proposed as a general phenomenon by \citem{BHMM}. They demonstrated this phenomenon in simple machine learning methods as well as simple neural networks. This phenomenon was observed for different models in Opper (1995; 2001), \citem{AS}, \citem{SGD}, and \citem{GSD}.  

Recently, \citem{NKB} show that double descent is a robust phenomenon that occurs in a variety of datasets, architectures, and optimization methods. In addition, they also demonstrate that the same double descent phenomenon can happen for training time (Òepoch-wise double descentÓ) and dataset size (Òsample-wise non-monotonicityÓ). One observes the epoch-wise double descent phenomenon for sufficiently large models. Holding model size fixed and training for longer also exhibits the double descent behavior. 

However, for medium-sized models, as a function of training time, the test error follows the classic U-shaped curve. For small models, the test error decreases as the training time increases. Oddly, the phenomenon may lead to a regime where more data hurts. The performance first improves, then gets worse, and then improves again with increasing data size. In summary, double descent occurs not just as a function of model size, but also as a function of training time and dataset size which means more data may surprisingly lead to worse test performance.

Based on the universal double descent phenomenon with respect to model size, training time or sample size, \citem{NKB} propose a new measure of complexity, the effective model complexity (EMC). EMC is related to other classical complexity measures such as Rademacher complexity and VC dimension. However, neither Rademacher complexity nor VC dimension can be applied to training time double descent effects since they depend on the model and the distribution of the data, but not on the training procedure. 
The effect of double descent phenomenon can be avoided through optimal regularization; e.g., \citem{NVK}.

The double descent is a universal phenomenon in terms of model size, training time and sample size. 
But complete understanding of the mechanisms and theory behind double descent in DL is an area of active research.

\subsection{Kullback-Leibler Divergence Reversal}\label{sec:KL}

Suppose (as with GANs), one wants to generate complex data that
resemble reality, such as fake images.
One has a sample of images $X_1, \ldots, X_n$ that is i.i.d.\ $G$, but the 
analyst has incorrectly specified the model family as $F_{\theta}$.
The analyst seeks to estimate $\theta$ by $\hat{\theta}$, in order to 
draw samples from 
$F_{\hat{\theta}}$ that seem realistic.

To a statistician, the natural approach is maximum likelihood estimation,
which gives
\begin{eqnarray}
\hat{\theta} &=& \mbox{argmax}_{\theta} n^{-1} \sum \ln f_{\theta}(x_i)\\
&\approx& \mbox{argmax}_{\theta} \E_G[ \ln f_{\theta}(x)].
\end{eqnarray}
where $f_{\theta}(x)$ is the density of $F_{\theta}(x).$
We note, for future use, that the maximizer does not change if one adds
the entropy term for $G$, so
\begin{eqnarray}
\hat{\theta} &\approx& \mbox{argmax}_{\theta} \E_G[ \ln f_{\theta}(x)] - 
\E_G \ln g(x) \\
&=& \E_G [\ln \frac{f_{\theta}(x)}{g(x)}] \\
&=& \mbox{KL}(G || F_{\theta})
\end{eqnarray}
where the last term is the Kullback-Leibler divergence from $F_{\theta}$ to
$G$.

In contrast, in DL, \citem{carin} reversed the roles of $G$ and $F_{\theta}$ 
and sought 
$$
\tilde{\theta} = \mbox{argmax}_{\theta} \E_{F_{\theta}} [ \ln g(x)].
$$
Obviously, this cannot be calculated, since the analyst does not know
$G$, but it turns out there is still a way forward.

Suppose that the support of $F_{\theta}$ is a proper subset of the 
support of $G$ (in a very high dimensional image space).
Then $G$ gives positive probability to pixel arrays that cannot
be generated from $F_{\hat{\theta}}$, and thus draws from $F_{\hat{\theta}}$
will likely not resemble draws from $G$.
But the role reversal done in DL ensures that the $\tilde{\theta}$
maximizer assigns positive probability to images or text generated
by $G$.
They will seem real, although they may not explore the full range
of realizations from $G$.

To encourage wider support, \citem{carin} imposed an entropy regularization,
obtaining
\begin{eqnarray}
\tilde{\tilde{\theta}} &=& \mbox{argmax}_{\theta} \E_{F_{\theta}} [ \ln g(x)]
- \E_{F_{\theta}}[\ln f_{\theta}(x)]\\
&=& \mbox{argmax}_{\theta} \E_{F_{\theta}} \ln \frac{g(x)}{f_{\theta}(x)} \\
&=& \mbox{argmin}_{\theta} \mbox{KL}(F_{\theta} || G)
\end{eqnarray}
which is the Kullback-Leibler divergence from $G$ to $F_{\theta}$.
The entropy term favors $F_{\theta}$ that have large support.

Since $F_{\theta}$ is unknown, the divergence cannot be calculated.
But let $v(x) = \ln g(x)/f_{\theta}(x)$
and suppose one is able to easily sample from a distribution $H(z)$, which
has an analytic form.
Then construct the model $x = w_{\theta}(z)$ where $w{\theta}(z)$ is a
deep network and $z$ is a latent variable (the Òreparameterization trick").
Now the problem is to find
$$
\tilde{\theta} = \mbox{argmax}_{\theta} \E_{H} [g(w_{\theta}(z))] 
$$
which brings $\theta$ inside the expectation.

Suppose one knew $g(x)$ and $f_{\theta}(x)$ and wanted to find the
optimal classifier to determine from which distribution an observation
came.   
The optimal classifier is just the log likelihood ratio $v(x)$.
Since one has samples from $g(x)$ (the data) and can sample
$f_{\theta}(x)$ through the latent variable $z$, then one can
construct a second deep network that does binary classification
to determine whether an observation was drawn from $f_{\theta}(x)$
or from $q(x)$.  
This trains the new deep network to learn $v(x)$.
By training both neural networks (the GAN framework), they converge 
to an equilibrium point at which the deep network corresponding
to $f_{\theta}(x)$ produces outputs (e.g., images) that look
like observations produced by $g(x).$

\subsection{Toward Uncertainty Quantification}

DL models are often viewed as deterministic functions, which limits
use in many applications.
First, model uncertainty cannot be assessed; statisticians know this can 
lead to poor prediction (\citem{gal2017deep}).
Second, most network architectures are designed through trial and error, or 
are based upon high level abstractions. 
Thus, the process of finding an optimal network architecture for the task at hand, given the training data, can be cumbersome.
Third, deep models have many parameters and thus require huge amounts of training data. Such data are not always available, and
so deep models are often hampered by overfitting.   

Potentially, these issues could be addressed using Bayesian statistics
for uncertainty quantification (UQ). 
UQ builds a Gaussian process emulator for a complex simulator, which is
one way to view DL networks.
That approximation enables probabilistic expression of uncertainty
(\citem{gramacy}).
It also enables fast but indirect study of architectures and training
regimens.
Random effects modeling makes the Bayesian emulator resistant to overfit.
Uncertainty quantification for deep learning has recently gained momentum
(\citem{zhu2019physics}, \citem{tripathy2018deep}). 
  
The current status of this research is summarized below. 
\begin{itemize}
	\item Bayesian modeling and variational inference. The true posterior $p(w|X, Y)$ cannot usually be evaluated analytically. Instead, a variational distribution can be evaluated to approximate the posterior distribution. Variational inference is a standard technique to replace the Bayesian modeling marginalisation with optimization. Compared to the optimization approaches used in DL, this method optimizes over distributions instead of point estimates. This approach preserves many of the advantages of Bayesian modeling (such as the balance between complex models and models that explain the data well), and results in probabilistic models that capture model uncertainty.  
	\item Bayesian neural networks.  These offer a probabilistic interpretation of DL models by inferring distributions over the models' weights. They offer robustness to over-fitting, uncertainty estimates, and can learn from small datasets. 
	\item Deep Gaussian Process (DGP) Methods.  The idea behind using DGP methods for non-stationary functions comes from the DL's approximation of complexity through composition---one can create very flexible functions from compositions of simpler ones. So DGP methods with functional compositions of GPs have been developed to construct probabilistic prediction models and handle forecasting uncertainties (\citem{lee2017deep}).   
\end{itemize}
These UQ methods enhance DL by enabling probabilistic statements of uncertainty
about DL classification decisions and predictions.

\subsection{Adversarial Learning}

It is important to recognize that DL can fail badly. In particular,
DL networks can be blindsided by small, imperceptible perturbations in the inputs,
also known as data poisoning (\citem{naveiro2019classification}, \citem{vorobeychikK18}).  The attackers need not even know the details of the machine learning model used by  defenders~(\citem{moosavi-dezfooli17FF}).

One important case is autonomous vehicles,
where DL networks decide, e.g., whether the latest sequence of images implies that the car should apply its brakes.  There are now many famous cases in which changing a few pixels in training data can creates holes in DL network performance (\citem{modas2019sparsefool}, \citem{su2019one}). Symmetrically, a small perturbation of an image, such as applying a post-it note to a stop sign, can fool the DL network into classifying it has a billboard and not recognizing that the vehicle should brake (\citem{eykholt_2018_CVPR}). 

\section{Detailed Syllabus and Example Homework}

The content of a DL course must depend upon the background of
the students, the goal of the department, and the needs of the
students.
The proposed syllabus assumes the target audience are PhD graduate
students in statistics, mathematics, and/or computer science.
It assumes
that proficiency in practice is wanted by both the department and 
the students, with a secondary goal of fostering some insight
into the theory.

These assumptions describe the situation encountered by one of the
authors in developing such a course.
The class was not part of any department's core program, but its
popularity signalled the appetite for such material.
Students were able to train deep networks on their laptops, but
it is important to create assignments that prescribe limits on
the amount of training that is done, partly to ensure comparability
during grading and partly to prevent overuse of computing 
resources.

In a DL class, there are challenges and choices both in terms of coverage
and evaluation.
Our sense is that for most cases (with the possible exception of short
courses) regular homework assignments that build skills in training 
deep networks is highly valued by the students.

\subsection{Syllabus}
The following syllabus and its rationale grew out of two graduate courses in DL; one was taught in the fall of 2019, and the other was taught in the
summer of 2020.
Our syllabus has evolved based on those experiences.

\begin{enumerate}
\item The first lecture should motivate the material.  It should begin by
listing some of the
successes of DL (e.g., Tesla's Autopilot, Google Translate, Alpha Go, its defeat
of the Dota2 champions), and also
cool applications such as style transfer, image reconstruction, image synthesis,
object segmentation, colorization, super-resolution, and caricature creation.

Next, one could show the improvement over time in the ImageNet classification
competition, and relate that to the depth of the DL network.
%Figure~\ref{ImageNet} illustrates this material.
Emphasize that DL requires lots of training data.

\item Define the perceptron and describe 
how it works.
Define the input, output and hidden layers---a visual image is obviously
helpful and ubiquitously available on the Internet.
Mention the affine transformation and the activation function, but not 
in detail---that will get covered in the next lectures.
If the audience are mostly statisticians, point out that in the DL world,
the constant in the affine transformation is usually called a bias, although
to us it would be an intercept.
Emphasize that a deep network creates complex outcomes by function 
composition rather than linear modeling, as statisticians have been 
trained to do.

Whether or not the class is going to do exercises (and we strongly recommend
that it should), it is important for students to know some of the
standard benchmark datasets.
We used the MNIST digit and fashion data,
the Cifar-10 data, and the Google Street View House Numbers data (these
are listed in order of increasing difficulty).
Later in the course, for text applications instead of images, 
we used the IMDB Movie Review data.
Both MNIST datasets, the Cifar10 data and the IMDB data are available
at \url{https://keras.io/api/datasets/}, and the
Streetview data are at \url{http://ufldl.stanford.edu/housenumbers/}.
One should not assign the students datasets that are too difficult---these
require lots of data, lots of time, and lots of electricity to train, which 
interferes with the educational objectives. 

\item Compare and contrast some of the standard loss functions and activation
functions.
But the main content for this lecture should be the backpropagation
algorithm.
Define the gradient, and show how the chain rule operates on
compositions of function to allow estimation of the coefficients in the
affine transformations in the activation functions.
Introduce stochastic gradient descent.

If the class is mathematically advanced, one can describe the Universal Approximation Theorem.
It reassures many statisticians that DL has theoretical underpinnings.

\item Generalize the previous lecture to tensors and discuss TPUs.
This is probably a good time to talk about the enormous number of
parameters in DL, and introduce ideas of sparsity, regularization, 
the bias-variance tradeoff and
the Curse of Dimensionality.
Discuss the surprising success of DL, given these considerations,
and mention double descent.

\item Introduce initialization, weight decay, minibatches, learning
rates and epochs.  Explain saturation and momentum.  

The following topics vary in their practicality, but if the course
has a theoretical component, they should be addressed: AdaGrad,
RMSProp, Nesterov momentum, second-order gradient methods,
conjugate gradient descent, block coordinate gradient descent, batch renormalization and Polyak averaging (approximately in this order).
This will probably take two lectures.

\item A statistical audience will want to know about the
Robbins-Monro algorithm (\citem{Robbins-Monro}) and how it relates to DL.
Even if the audience is non-statistical, a sketch of the algorithm
is valuable.  
However, if pressed for time or focused on applications much more
than theory, this topic is dispensable.

\item At some point, we recommend stepping back from the
mathematics and comparing the designed experiments that are
embedded into the homework assignments (see subsection 3.2).
At this point, the class has probably completed three or four
assignments on the MNIST fashion data and/or the Cifar10 data.
The experiments can compare the effect of activation functions,
learning rates, breadth versus depth, and minibatch size on
performance.

\item  We recommend revisiting regularization techniques.
In DL, these include weight decay, L1 penalties, quantization,
quantization through k-means clustering, and quantization through
Huffman coding.
Discuss data augmentation, noise robustness, label smoothing, 
the softmax function, early stopping, bagging, stacking, and dropout.
This will take two to three lectures.

If one needs to fill out the third lecture, one can introduce the
computation graph. 
But this topic may also arise later.

\item Now do a deep dive on convolutional neural networks (CNNs).
Introduce the convolution operation, talk about sparse interactions,
parameter sharing, zero (and smarter) padding, and pooling.
Discuss convolution with stride and tiled convolution.

It is probably good to remind students of the successful applications of 
CNNs mentioned in the first lecture. 
This is a good motivator for a lecture on reinforcement learning.

Showing some simple computation graphs is helpful, and will feed
forward into the next lesson.
Covering this material will take about 2.5 hours worth of
lecture time.

\item Next come Recurrent Neural Networks (RNNs).
Discuss the standard architectures, and introduce backpropagation through time.
Computation graphs, and other visuals, are essential.
Describe teacher forcing, skip connections, and leaky units.
Then proceed to long short-term memory, gated recurrent units, the constant
error carousel, augmented RNNs (including neural Turing machines),
and echo state networks.

RNNs are prone to vanishing and exploding gradients, so although that 
also can arise in CNNs, we recommend covering it here.
Mention gradient clipping for exploding gradients.
In our experience, this is about 1.5 hours of material.

\item
RNNs are dominant when dealing with text or language,
so the homework on this unit should employ the IMDB database.
When covering text examples, and it is 
helpful to describe Word2Vec, Seq2Seg, and Google's BERT.
The concept of attention arises naturally.

Recursive Neural Networks (RvNNs) are not as prominent as other
DL techniques, but a short description is appropriate here. 

\item Variational autoencoding is the next topic.
This can be motivated to a statistical audience by pointing out that
a shallow network with a single hidden layer and linear activation
functions is equivalent to principal components analysis.
Describe the denoising autoencoder.
There is hardly any new theory in this unit, but there are a
number of compelling applications.

For the homework in this segment, we recommend that the
class assignments compare information loss across a range
of architectures for simulated data in which the signal-to-noise
ratios are known.  
Insofar as we are aware, little is known about how to pick
the right autoencoding architecture to match an application
with specific characteristics.

\item The last main unit concerns Generative Adversarial 
Networks (GANs).
After introducing the key ideas and the game-theoretic
underpinning, one can proceed to conditional GANs, 
ancestral sampling, and diffusion inversion.
Emphasize that the GANs are learning a low-dimensional 
manifold on which the images (or other material, such as
text) concentrate.
Wasserstein GANs (WGANs) should be mentioned and perhaps covered
in some detail.
WGANs show more stability during training than regular
GANs and use a loss function that better reflects image quality (\citem{gulranjani}).

For a statistical audience, the work in \citem{carin}
nicely connects ideas in maximum likelihood estimation to GANs.
Since it is relatively new, Section \ref{sec:KL} reviews it. 

\end{enumerate}

This syllabus lays out one path through DL material, emphasizing connections to
statistical thinking.
If time permits, the instructor may want to include more material adjacent to
DL, such as empirical risk bounds, or other statistical
approaches to classification and text data.

\subsection{Homework}

If all students are given the same assignment, say to train a deep network
to classify the Google streetview data, and if all are using the same
programming environment, then one must be very prescriptive to
ensure that the results are comparable.
The Tesla automated driver DL system took 70,000 GPU hours to train
(see Autopilot AI, \url{https://www.tesla.com/autopilotAI}), and we do not want students to be marked on the basis of how long they let their programs run.
So one must specify the number of training epochs, the architecture of their
networks, and many other details.

The MNIST digit data is quite easy to classify well, so it is a good
starting point for homework. 
But later assignments should be more difficult
in order to illustrate realistic DL performance.

We found it interesting to set up the homework as a designed experiment.
Specifically, for a large class, we recommend that triplets of students be
given the same assignment, and the assignments should vary in ways that study some
of the DL strategies.
This enables an ANOVA with three replicates for each combination of factor
levels.

For example, if the students are not all using the same programming
environment, then the environment might be one of the design factors, so students
can compare, say, PyTorch to TensorFlow to Keras.
Similar designed experiments can study the impact of different architectures,
learning rates, and dropout protocols.
Such experiments can help students understand the impact of different 
decisions that are typically made when training a DL network.
(Note:  Typically the results of such assignments will be error rates,
which are binomial random variables; therefore, we recommend using the
arcsine square-root transformation before fitting the ANOVA.)

In our class, the first project was to fit a fully connected feedforward
network to the MNIST digit data.
There were to be 3 hidden layers, with each layer having 512 nodes.
Students whose ID number ended in a 0 or 1 (or 2, 3, or 4, 5 or 6, 7 or 8, 9)
used the ReLU activation function (or sigmoidal, tanh, linear, or
leaky ReLU, respectively), and students whose
terminal ID digit was even used minibatchs of size 100 for two epochs
and those that were
odd used minibatch sizes 200 for one epoch (so the total training effort is
equivalent).  The loss function was Kullback-Leibler divergence.
Students reported results classification accuracy and training time.

As a slightly more advanced example, one might direct students
to fit a fully connected feedforward network with 4 hidden layers,
where each hidden layer has 1,024 nodes, to the Google streetview data.
Students whose ID number ends in a 0 or 1 (or 2, 3, or 4, 5 or 6, 7 or 8, 9)
should use a learning rate of 0.8 (or 0.85, 0.9, 0.95, or 0.99, respectively), and students whose
terminal ID digit is even use the cross-entropy loss 
and those that are odd use 0-1 loss.  All minibatches should be of size 200 and the
network should be trained for ten epochs.
Students should report classification accuracy and training time.

In general, even more prescriptive detail will be needed to ensure reasonable
comparability among the graded homework.

%%%%%%%%%%%%%%%%%%%%%%%%%%%%%%%%%%%%%%%%%%%%%%%%%%%%%%%%%%%%%%

\section{Review of DL Resources}
Numerous DL resources and courses have been developed and
are available online. 
The question is where to start. 
We summarize the most popular DL resources to provide a learning path 
for students who are new to DL, and also to those who want to explore it further. 

\subsection{Online Courses}
There are many DL courses and tutorials available online. 
The most popular ones include: 
\begin{itemize}
	\item The Stanford University Online Open Courses.
	\begin{itemize} 
	\item CS230: Deep Learning (\url{https://cs230.stanford.edu/})
	\item CS231n: Convolutional Neural Networks for Visual Recognition (\url{http://cs231n.stanford.edu/})
	\item CS224d: Deep Learning for Natural Language Processing \\(\url{http://web.stanford.edu/class/cs224n/}. 
	\end{itemize}
	This series of courses was compiled by deep learning experts, such as Fei-Fei Li,
	Andrej Karpathy, Samy Bengio, Tom Dean, Andrew Ng, and Richard Socher. The courses offer video explanations of topics including logistic regression and regularization of the cost function, vector implementation, polynomial regression, all from a DL perspective. These courses are designed for DL beginners. 
	
	\item A Deep learning specialization course series by Andrew Ng on Coursera \\(\url{https://www.deeplearning.ai/deep-learning-specialization/}). This is a suite of five courses on DL, starting with starter course and covering all major types of DL networks. This series provides an excellent in-depth introduction for beginners to DL models.  
	There are exercises on how to use these tools.
	   	
	\item The Deep Learning Nanodegree by Udacity. It is a popular training course which is designed in collaboration with leading AI experts \\(\url{https://www.udacity.com/course/deep-learning-nanodegree--nd101}). This course is offered by a team of instructors who are AI engineers and industry experts with a notable emphasis on hands-on projects and learning by doing. Each student will need to complete a portfolio of five real-world DL projects during the course to earn a nanodegree certification. The course also provides interactive mentor feedback to assist student learning. The course is not free, but it is designed for students who want to improve practical problem-solving skills and learn hands-on experience that prepares them to work in AI tech companies. 
		
	\item DataCamp: Interactive Learning Course (https://learn.datacamp.com/) is an emerging platform that provides interactive training courses on data science and programming. All the courses are designed to let students learn by doing.
	They apply each lesson immediately and get instant feedback. The DL courses target beginners and teach how to implement different neural network models quickly in an interactive learning environment with real-time feedback.   
	
	\item Simplilearn's Deep Learning Course (with Keras \& TensorFlow) Certification Training \\(\url{https://www.simplilearn.com/deep-learning-course-with-tensorflow-training}) is 
	a popular platform to teach DL.
	It is led by experienced trainers who use many real-life examples. 
	An important part of this teaching platform is that it provides well-organized prerequisite courses. 
	%%Once a student completes those, the student to can advance to DL courses.
	  
\end{itemize}

\subsection{Books}
There are many DL books. Some of the best include: 

\begin{itemize}
	\item {\em Deep Learning} by \citem{gb2016}. 
	It is aimed at people who have no knowledge of deep learning, but are mathematically sophisticated. It introduces DL methods and concepts, as well as advanced DL topics, including research directions and practical applications. 
   
    \item {\em Deep Learning with Python} by \citem{Francois2017}. 
    This book is written by the inventor of Keras inventor and introduces DL using Python and the Keras library. It balances theory and coding with intuitive explanations and practical examples.
    
    \item {\em Hands-On Machine Learning with Scikit-Learn and TensorFlow} by \citem{Aurelien2017}. this is a popular book that first employs Scikit-Learn to introduce fundamental machine learning tasks, and employs Keras and TensorFlow to learn deep neural networks. It uses examples to build understanding of the concepts and tools, and has practice exercises in each chapter.     

    \item {\em Deep Learning Tutorial} was developed by the LISA lab at University of Montreal. It is a concise tutorial. The book \url{http://deeplearning.net/tutorial/deeplearning.pdf} emphasizes use of the Theano library for building DL models in Python.  See \url{https://github.com/lisa-lab/DeepLearningTutorials}.
    
    \item {\em Neural Network Design} (2nd Edition) by \citem{Hagan2014}: This book provides a clear and detailed survey of basic neural network architectures and learning rules. The authors emphasize the mathematical analysis of networks, methods for training networks, and application of networks to practical engineering problems in pattern recognition, signal processing, and control systems.
\end{itemize}

\subsection{Popular Online Resources}
In addition to courses and books, there are also resources available online to learn about DL, real world applications, and recent advances. 
Some of the most active online resources are:   

\begin{itemize}
    \item Github: There are many educational resources and open source projects on Github. One can access many machine learning and deep learning projects and example implementations in Python and other programming languages. See  \url{https://github.com/topics/deep-learning/}.   
    
	\item Quora: A great resource for AI and machine learning. Many of the top researchers answer questions on the site \url{https://www.quora.com/} and it provides an active platform to share knowledge of DL-related topics. Under the Quora Deep Learning feed \url{https://www.quora.com/topic/Deep-Learning/}, one can check a curated list of questions discussed by the community.

    \item Reddit: There is an active DL community on Reddit. One can follow subreddits on machine learning \\ \url{https://www.reddit.com/r/MachineLearning/} and Deep Learning \\
    \url{https://www.reddit.com/r/deeplearning/} to keep up with the latest news and research. 
    
    \item Blogs: There are a number of active blogs that post consistently on AI-related topics with original materials. The top viewed ones include DeepMind Blog \url{https://deepmind.com/blog}, OpenAI Blog \url{https://openai.com/blog/}, Machine Learning Mastery  \url{https://machinelearningmastery.com/}, distill.pub (a modern medium for presenting research), Data Science Central  \\
    \url{https://www.datasciencecentral.com/}, Towards Data Science \url{https://towardsdatascience.com/}.   
	
\end{itemize}

\subsection{Developing Programming Skills}\label{sec:programming-skills}
Python will remain a dominant programming language in DL, and Jupyter
notebooks simplify many tasks.
But students should expect to learn many new languages and environments 
and tools over the course of their careers.
Statisticians are most comfortable with \textsf{R} and SAS,
but these are not much used in DL.
And the traditional statistical curriculum does not usually teach 
DL programming, for many reasons.

As a result, most students will need to teach themselves the 
programming environments they need to master.
Fortunately, there are on-line options:
\begin{itemize}
	\item CheckiO is a gamified website containing programming tasks that can be solved in Python 3.
    \item Codecademy provides lessons on Python.
    \item Code The Blocks combines Python programming with a 3D environment where one places blocks to construct structures. It also comes with Python tutorials that teach one to create progressively elaborate 3D structures.
    \item Computer Science Circles has 30 lessons, 100 exercises, and a message system where one can ask for help. Teachers can use it with their students. It is available in English, Dutch, French, German, and Lithuanian.
    \item DataCamp Python Tutorial is unlike most other Python tutorials.  It lasts
    four hours and focuses on Python for Data Science. It has 57 interactive exercises and 11 videos.
    \item Finxter tests and trains with 300 hand-picked Python puzzles.
    \item How to Think Like a Computer Scientist: Interactive Edition is a re-imagination of the \citem{EDM} Elkbook with visualizations and audio explanations. 
\end{itemize}

For students with little programming experience, several books can be useful.
These books can be purchased online, but also have free electronic versions. 
\begin{itemize}
	\item {\em Automate the Boring Stuff with Python: Practical Programming for Total Beginners} by \citem{sweigart1} is written for office workers, students, administrators, and anyone who uses a computer to learn how to code small, practical programs.
	\item {\em How To Think Like a Computer Scientist} by \citem{EDM} is a classic open-source Python book.  It was updated to Python 3 by Peter Wentworth.
\end{itemize}

\section{Real World Application Areas and Examples}

We describe two interdisciplinary application areas (medicine
and astronomy) that use DL but which have deep statistical roots. We encourage DL courses to motivate students with real world examples and projects. We further urge traditional statistical consulting classes to be re-engineered to include DL consulting experiences, of the kind described in the following examples.

In medicine, previous landmark studies have addressed a variety of classification tasks to detect patterns from image data sets using CNNs (\citem{Ardia2019}, \citem{Coudray2018}, \citem{Esteva2017}, \citem{Ehteshami2017}, \citem{Gulshan2016}) or an 
ensemble of CNNs (\citem{McKinney2020}). 
The studies apply DL to the various medical image data sets for diagnostic or screening purposes. 
Students who know traditional classification methods in statistics
can compare the performances on these tasks.

Recent work uses DL to predict patients' prognoses, including survival, recurrence, hospitalization, and treatment complications. 
Studies have built DL-based predictive models using RNNs (\citem{Rajkomar2018}, \citem{Tomasev2019}) or feedforward networks with fully connected layers (\citem{Lee2018}). 
Long short-term memory architecture has been used to automate natural language processing to extract clinically relevant information from electronic health records (\citem{Liang2019}). Probabilistic prediction and quantification of uncertainties are crucial topics to address, but the DL community spends
less attention on this aspect than statisticians do.

Patients' electronic health records are usually longitudinal data with 
time-to-event endpoints. To statisticians, understanding the nature of the data generation mechanisms and statistical modeling is key. 
Missing data problems arise frequently in these records. 
Strategies to deal with missing data need to be tailored to the specific
problem. 
But in DL, missing data are handled in rudimentary ways. 

In astronomy, DL has become popular with exoplanet detection, and detection of gravitational waves using LIGO-Virgo data. 
DL will be widely used by the Rubin Observatory Legacy Survey of Space and Time (LSST).

NASAÕs Kepler mission collected over 150,000 light curves on brightness of stars in our galaxy with the aim of detecting planets outside our solar system.  However, Earth-sized planets orbiting Sun-like
stars are on the very edge of the missionÕs detection sensitivity.  
DL is being used to classify potential 
planet signals (\citem{shallue and vanderburg}) and thus determine the frequency of Earth-sized planets.  
CNNs are trained to predict whether a given signal is an exoplanet or a false positive caused by astrophysical or instrumental phenomena.  

A new era dawned in 2015 with the detection of Gravitational
Waves (GWs). The first detection of GW150914 by the Laser Interferometer Gravitational-Wave
Observatory (LIGO) was from the merger of two black holes, each about 30 times the mass of our sun.  
But such events are rare, and DL is data hungry,
so the majority of the DL work in this area is based on simulated data.
\citem{george and huerta} take advantage of scientific simulations
to produce the data for needed for training. 
They focus on labeled data obtained from physics simulations 
(with domain knowledge) to train CNNs to detect signals embedded in noise and also estimate
multiple parameters of the source.
\citem{li et al 2020} use DL to detect GW events based on simulated signals contaminated with white Gaussian noise. It is a classification problem, 
and the method is particularly valuable when the potential GW signal shapes are too complex to be characterized.
Statistical knowledge is essential in creating faithful simulations.

The forthcoming LSST should detect $\sim$300M quasars out of the $\sim$40B objects in the all-sky survey, but experts expect to confidently identify only about 10\% of them. This is in contrast to the Sloan Digital Sky Survey (SDSS) dataset where about 98\% can be identified. 
While the true situation is somewhere between 10\% and 98\%, the cause of the problem is that the training set for quasars is based on bright objects 
%(roughly $<20$ mag)
while LSST will see much fainter objects. If fainter quasars were to reside in the same location as brighter quasars in the 4-dimensional SDSS color space, then the bright training set would work. But this is not true: fainter quasars are typically more distant with higher redshift, causing different emission lines to enter/exit different bands; and quasars probably have intrinsically different properties both at earlier cosmic times (called `cosmic evolution') and at lower luminosities (where host galaxy contamination becomes important). 

This problem arises in many applications, where the training sets will be based on certain samples which, if used with conventional classification procedures, will recover only biased subsamples of the members of the class. This 
kind of bias is well-understood by statisticians and should be taught
in DL courses.

\section{Final Remarks}
It is imperative that statisticians be involved in DL.
We can learn a lot, but also contribute much.
However, involvement requires us to learn and teach this material,
which poses practical, pedagogic and technical challenges.

This paper proposes partial solutions to many of those challenges.
We hope it will help teachers who are trying to extend
their curricula, and students who are eager to learn
this new field.

Teaching DL requires us to find new ways to think and to 
deliver educational content.  
We believe that the statistics profession is up for the 
challenge.

\section*{Acknowledgement and Funding Information} 
We thank deep learning working group at the Statistical and Applied Mathematical Sciences Institute (SAMSI), USA for helpful discussions. This work was supported by National Science Foundation grant DMS-1638521. Hyunsoon Cho's work was partially supported by the National Research Foundation of Korea grant (No. 2020R1A2C1A01011584) funded by the Korea Ministry of Science and ICT. Hailin Sang's work was partially supported by the Simons Foundation Grant (No. 586789). Shouyi Wang's work was partially supported by the National Science Foundation grant ECCS-1938895.

%%%%%%%%%%%%%%%%%%%%%%%%%%%%%%%%%%%%%%%%

%%%%%%%%%%%%%%%%%%%%%%%%%%%%%%%%%%%%%%%%%%%%%%%%%%%%%%%

%\bibliographystyle{authordate1}

\end{document}